\documentclass[sigconf, screen]{acmart}
\AtBeginDocument{%
  }

\setcopyright{acmlicensed}
\copyrightyear{2018}
\acmYear{2018}
\acmDOI{XXXXXXX.XXXXXXX}
\acmConference[Conference acronym 'XX]{Make sure to enter the correct
  conference title from your rights confirmation email}{June 03--05,
  2018}{Woodstock, NY}
\acmISBN{978-1-4503-XXXX-X/2018/06}

\usepackage{pifont}
\begin{document}

\title{DeepSight: Bridging Depth Maps and Language with a Depth-Driven Multimodal Model}
% \bibliography{name}
\author{Hao Yang}
\authornote{Both authors contributed equally to this research.}
\email{hyang@ir.hit.edu.cn}
\orcid{0000-0002-8138-7387}
\author{Honbo Zhang}
\authornotemark[1]
\email{hbzhang@ir.hit.edu.cn}
\affiliation{%
  \institution{Harbin Institute of Technology}
  \city{Harbin}
  \country{China}
}

\author{Yanyan Zhao}
\affiliation{%
  \institution{Harbin Institute of Technology}
  \city{Harbin}
  \country{China}}
\email{yyzhao@ir.hit.edu.cn}

\author{Bing Qin}
\affiliation{%
  \institution{Harbin Institute of Technology}
  \city{Harbin}
  \country{China}}
\email{qinb@ir.hit.edu.cn}

% \affiliation{%

% \affiliation{%

% \affiliation{%

% \affiliation{%

% \affiliation{%

\renewcommand{\shortauthors}{Trovato et al.}

\begin{abstract}
Multimodal large language models (MLLMs) have achieved impressive performance across various tasks such as image captioning and visual question answer(VQA); however, they often struggle to accurately interpret depth information inherent in visual data. In this work, we introduce DeepSight, the first dedicated depth MLLM designed to enhance three-dimensional scene understanding. Unlike conventional methods that align RGB image encodings with text, our approach takes advantage of the unique characteristics of depth images: single-channel grayscale images where the pixel values directly reflect depth cues to improve spatial reasoning. To address challenges associated with limited depth data and the inadequacy of simple channel replication, we construct a novel depth image-text pair dataset and a depth instruction dataset. Depth maps are generated from visual images using the GLPN model, and GPT-4 is employed to curate corresponding depth instructions, an approach validated by LLaVA. Additionally, we modify the ViT encoder in CLIP to incorporate local object information, thereby capturing the subtle continuous variations of depth more effectively. To evaluate the performance of our model, we develop a comprehensive depth question answer benchmark based on existing depth image datasets, which rigorously assesses understanding in typical depth map scenarios. Experimental results demonstrate that DeepSight significantly enhances depth perception and downstream task performance, marking a substantial step forward in multimodal three-dimensional understanding.
\end{abstract}







\maketitle

\begin{figure}[htp]
    \centering
    \includegraphics[width=8.5cm]{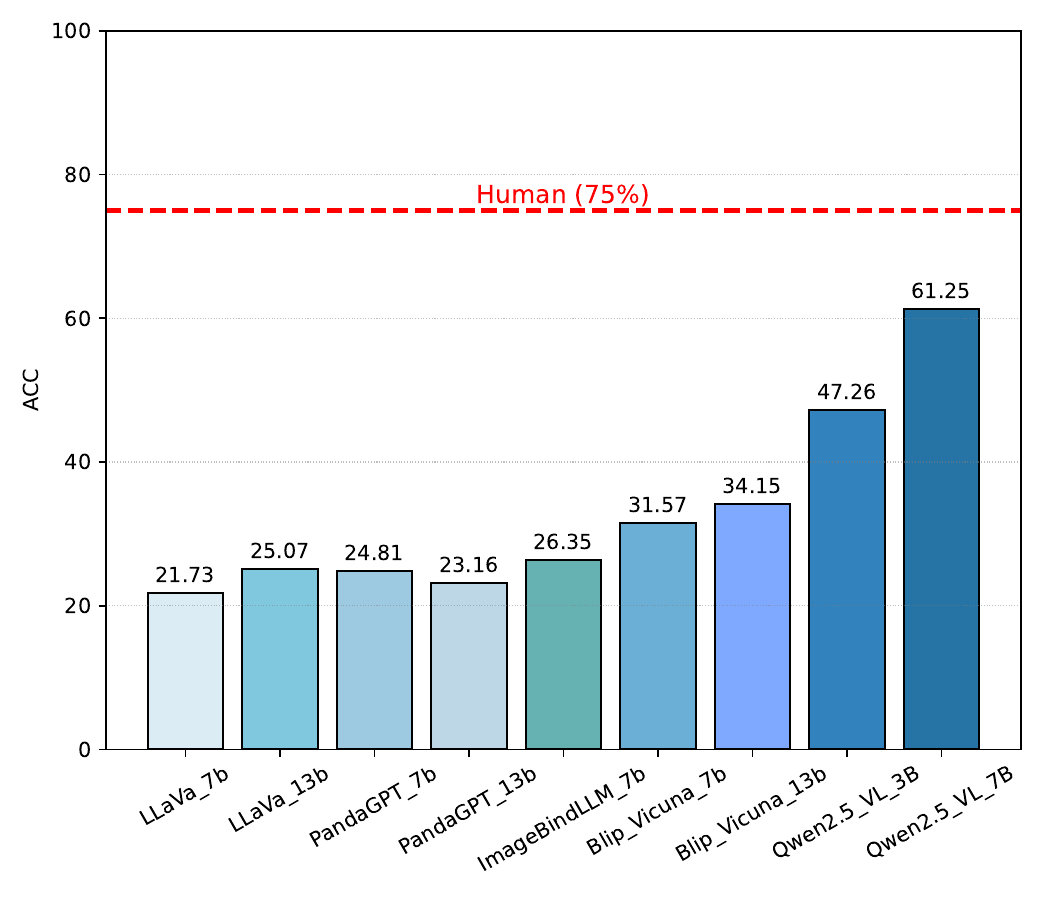}
    \caption{\textbf{Stereoscopic Vision Ability Test for MLLMs.} We input RGB images into the MLLMs and ask questions about distance comparisons between objects. The results show that the MLLMs have issues with its stereoscopic vision ability.}
    \label{fig:example}
\end{figure}

\section{Introduction}
Large Language Models (LLMs) have shown remarkable performance on numerous natural language processing tasks~\cite{touvron2023llama}. Building on this foundation, Multimodal Large Language Models (MLLMs) further extend this capability to vision-related downstream tasks such as image captioning and image-text retrieval~\cite{li2023blip2bootstrappinglanguageimagepretraining,Berkeley_Cmu_San,bai2023qwentechnicalreport,chan2023chatevalbetterllmbasedevaluators,Dai_Li_Li_Huat_Zhao_Wang_Li_Fung_Hoi,peng2023kosmos}, achieving impressive results. However, our experiments reveal that existing MLLMs still struggle to accurately interpret distance information within images.

As illustrated in Figure~\ref{fig:example}, we conduct a Stereoscopic Vision Ability Test by providing MLLMs only with RGB images and asking them to compare the distances between objects. The results reveal that MLLMs frequently misjudge which object is closer, highlighting a clear limitation in their spatial understanding. To enable MLLMs to have human-like stereoscopic vision, we believe that incorporating depth information is crucial. Yet, limited data availability and insufficient research on depth-based techniques for MLLMs remain a challenge.

Depth images typically consist of a single grayscale channel whose pixel values explicitly convey spatial distance. Compared with RGB images, depth information can significantly benefit the model’s grasp of three-dimensional structure. By aligning the depth encoding space directly with textual encodings, rather than solely with RGB encoders, MLLMs can achieve a more comprehensive understanding of 3D environments and thus perform better on downstream tasks. Including depth supervision during fine-tuning enables MLLMs to accept depth images as input and yield more accurate results on depth-related downstream tasks.

Previous work such as ImageBind~\cite{girdhar2023imagebind} has unified multiple modality encoding spaces by aligning them with RGB image encoders, using only RGB-text pairs for both alignment and instruction tuning. While it offers an initial approach to depth-image comprehension, ImageBind does not fine-tune its depth encoder, thereby limiting the model’s ability to process depth information fully. Hence, constructing dedicated depth image-text pairs and depth instruction data is crucial for enhancing the depth comprehension of such models.

At the same time, to mitigate the scarcity of depth data, we use GLPN~\cite{kim2022globallocalpathnetworksmonocular} to convert RGB images into depth images and employ GPT-4~\cite{ye2023comprehensivecapabilityanalysisgpt3,kalyan2024survey} to generate additional depth instruction data for fine-tuning~\cite{liu2024improvedbaselinesvisualinstruction}. Similar approaches have proven effective in previous work like LLaVA~\cite{liu2023visualinstructiontuning}. Finally, to systematically evaluate MLLMs’ depth comprehension, we construct a depth question-answering benchmark comprising several sub-tasks based on common depth-related scenarios. This benchmark accurately measures three-dimensional reasoning through a series of template-driven evaluations. In summary, our contributions are as follows:

\begin{itemize}
\item We introduce a dedicated benchmark for assessing how well models understand depth information in real-world scenarios. Built on existing depth datasets, this is the first systematic attempt to measure MLLMs’ stereoscopic vision ability through multiple, representative sub-tasks.
\item 
We enhance the ViT within CLIP by incorporating local object information as additional inputs, aiming to improve the model’s capacity to capture fine-grained details and interpret spatial relationships. This adjustment is particularly beneficial for tasks that demand precise depth perception and object interaction.
\item 
We present DeepSight, the first multimodal LLM specifically designed to integrate depth data with text. By incorporating generated depth data in both the alignment and supervised fine-tuning stages, we strengthen the model’s ability to correlate different modalities and refine its understanding of depth images, resulting in more accurate 3D perception.
\end{itemize}

\section{Related Work}

\textbf{Multimodal Large Language Models.}
In recent years, multimodal large language models (MLLMs) have made significant progress, achieving impressive results across a variety of tasks involving both visual and textual understanding. A central theme in this line of work is the alignment between visual encoders and large-scale language models, enabling the unified processing of image-text pairs~\cite{li2023blip,zhang2023video,li2022blip,lin2024vila}. For instance, MiniGPT-4~\cite{zhu2023minigpt4enhancingvisionlanguageunderstanding,chen2023minigpt} integrates a ViT-based vision encoder with the Vicuna~\cite{Berkeley_Cmu_San,touvron2023llamaopenefficientfoundation} language model through a linear projection layer, demonstrating strong image captioning and reasoning abilities. LLaVA~\cite{liu2023visualinstructiontuning} similarly connects the pre-trained CLIP visual encoder (ViT-L/14) to Vicuna and further boosts performance through instruction tuning.
In contrast to these direct alignment methods, our approach adopts a two-stage training paradigm: we first optimize the visual and language components independently, then fine-tune them jointly in a task-specific SFT stage. This design improves alignment while preserving the strengths of each modality.

\textbf{Depth Images.}
Depth images are a specialized form of visual data in which each pixel represents the distance from the camera to surfaces in the scene~\cite{Cao_Wu_Shen_2018,jiang2018rednet,hu2019acnet,hazirbas2016fusenet}. Unlike standard RGB images that capture texture and color, depth images emphasize geometric structure, often represented as grayscale intensity—closer objects appear brighter, while distant objects are darker. These images are typically obtained via stereo vision, time-of-flight sensors, or structured light, and are widely applied in robotics, 3D reconstruction, and autonomous driving~\cite{cai2024spatialbot}.
In computer vision, depth information enhances spatial understanding, enabling more robust object detection, pose estimation, and obstacle avoidance. Despite their value, depth images pose unique challenges due to their lower texture richness and domain differences from natural images. Our work aims to better integrate this underutilized modality into the MLLM paradigm, enhancing multimodal reasoning with 3D-aware representations.

\textbf{Scarce-Modality Large Language Models.}
While most MLLMs are developed for widely available modalities such as RGB images, audio, and video, less attention has been paid to rare or underrepresented modalities like depth, thermal, or radar~\cite{li2023videochat}. Works such as ImageBind~\cite{girdhar2023imagebind}, LanguageBind~\cite{zhu2024languagebindextendingvideolanguagepretraining}, and PandaGPT~\cite{su2023pandagptmodelinstructionfollow} aim to support diverse modalities by projecting them into a shared embedding space. However, these models typically rely on massive heterogeneous datasets and are not specifically optimized for scarce modalities.
Our approach diverges by focusing specifically on depth as a primary visual input, and constructing a dedicated depth-text paired dataset for alignment. Through a two-stage training framework—alignment followed by instruction tuning—we achieve efficient learning of depth semantics and effective cross-modal reasoning. This targeted strategy not only reduces resource demands but also improves depth-specific performance in tasks such as distance comparison, object recognition, and spatial reasoning.

\section{Depth Template Benchmark}
\label{Depth Template Benchmark}

\begin{figure}[ht]
    \centering
    \includegraphics[width=\linewidth]{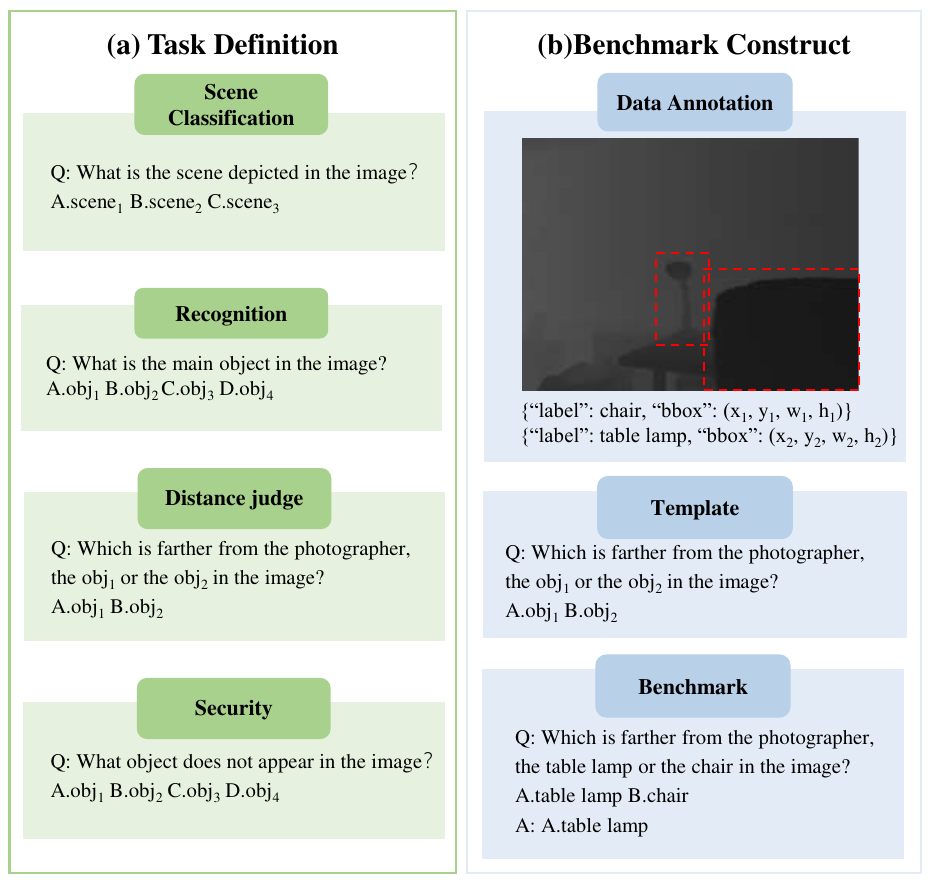}
    \caption{\textbf{Benchmark construction.} (a) defines the types of tasks. (b) describes how benchmark is constructed from templates.}
    \label{fig:2}
\end{figure}

Existing evaluation methods for depth modality remain relatively limited and simplistic, with most prior work focusing primarily on scene classification task~\cite{brown2011multi}. However, this narrow scope fails to provide a comprehensive assessment of a model’s capability to understand depth information. One key limitation in the field is the lack of standardized and fine-grained evaluation benchmarks specifically tailored to the depth modality.
To address this gap, we propose a depth-based question answering benchmark designed to assess MLLMs' depth perception and reasoning capabilities~\cite{silberman2012indoor,7298655}. All depth images used in our benchmark are real-world samples containing rich semantic and spatial information, enabling a more thorough and meaningful evaluation of the model's perceptual understanding. By introducing this benchmark, we aim to establish a more rigorous and targeted evaluation protocol for underexplored modalities like depth.

\subsection{Task Definition}
We define four sub-tasks: Scene Classification, Recognition, Distance Judge, and Security. These tasks are specifically designed to assess both the model’s holistic understanding of the overall scene and its ability to reason about fine-grained, object-level depth cues within depth images. Each task is formulated in a question-answer format, where the model either selects the most suitable answer from a set of candidates or generates a descriptive caption based on visual input.

\textbf{Scene Classification.}
This task evaluates the model's global perception by requiring it to classify the entire scene from a depth image. Given three candidate labels, the model must choose the one that best describes the environment. Success in this task reflects an understanding of spatial composition, layout, and contextual depth features that define different scene categories.

\textbf{Recognition.}
In this task, the model is prompted to identify specific objects located in targeted regions of the depth image. With four candidate answers provided, it must select the label that correctly matches the object. This evaluates the model's ability to extract meaningful object-level features from geometric cues present in depth data.

\textbf{Distance Judge.}
This task tests the model’s depth reasoning by asking it to compare the relative distances of two specified objects. Based on region-wise average depth values, the model must determine which object is farther from the viewpoint. This sub-task focuses on the model’s ability to interpret spatial hierarchy and construct a reliable depth-based relational understanding.

\textbf{Security.}
This task evaluates the model’s ability to judge the completeness of object recognition in a scene. Given four candidate objects, the model is required to identify the one that does not appear in the image. This task focuses on the model’s robustness in maintaining reliable recognition performance under complex scenes and potential category confusion.

\subsection{Benchmark Construct}

As shown in Figure~\ref{fig:2}, we developed the Depth Template Benchmark by constructing multiple problem templates and utilizing real depth images along with object bounding box annotations. We extract object instance segmentation from the bounding boxes to obtain the precise pixel regions of each object.The average depth within these regions is then utilized to estimate the depth of each object, thereby enabling a relative Distance Judge task among objects.

The distribution of the four sub-task categories is shown in Table~\ref{tab:task_num}, with a total of 13,473 question-answer pairs generated.

\begin{table}[ht]
  \caption{Task statistics across datasets. We construct four categories of sub-tasks: Scene Classification, Recognition, Distance Judgment, and Security. In total, 13,473 question-answer pairs are generated.}
  \label{tab:task_num}
  \centering
  \begin{tabular}{lccc}
    \toprule
    \textbf{Task} & \textbf{Source} & \textbf{Count} & \textbf{Percentage (\%)} \\
    \midrule
    Scene Classification  & NYU-D & 1,786 & 13.26 \\
    Recognition           & SUN-D & 3,793 & 28.15 \\
    Distance Judgment     & SUN-D & 5,737 & 42.58 \\
    Security              & SUN-D & 2,157 & 16.01 \\
    \midrule
    Total                 &        &13,473 &100.00 \\
    \bottomrule
  \end{tabular}
  \Description{A table showing the number of tasks across datasets: NYU-D and SUN-D. Tasks include scene classification, recognition, distance judgment, and security.}
\end{table}

\section{Depth Instruction Dataset}
\label{Depth Instruction Dataset}

\begin{figure}[ht]
    \centering
    \includegraphics[width=\linewidth]{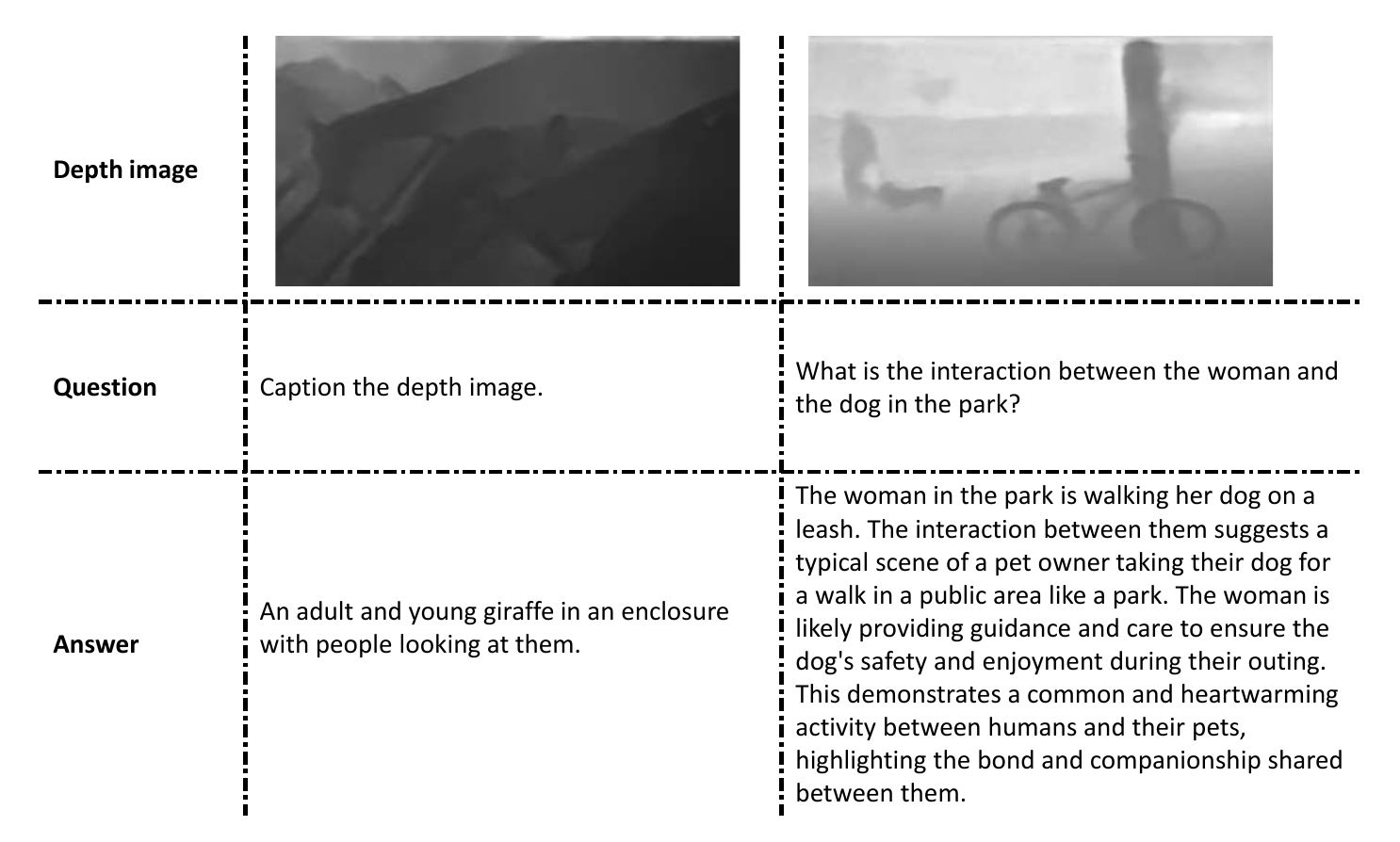}
    \caption{\textbf{Sample from the Depth Instruction Dataset.} The left side shows a sample of caption data used for alignment, and the right side shows an instruction sample generated by GPT-4.}
    \label{dataset exmaple}
\end{figure}

In this section, we describe the process of constructing the Depth Instruction Dataset. Due to the small scale of real-world depth datasets, which is insufficient for pretraining, we chose to translate RGB images from the COCO ~\cite{lin2015microsoftcococommonobjects} dataset into depth images. The dataset construction pipeline consists of three key steps: image translation, caption scoring, and instruction generation. Ultimately, we obtained 118K depth-text-bboxes pairs and 22k depth instructions as shown in Fingure ~\ref{dataset exmaple}.

\textbf{Image Translation.} We employ the GLPN model to transform RGB images into depth images. The resulting depth images preserve partial semantic information from the original RGB images while gaining enhanced spatial perception, facilitating a more comprehensive understanding of scene geometry. Although the depth information in these generated images is not ground-truth being predicted rather than directly measured our experiments demonstrate that this approach remains highly effective.

\begin{figure*}[t]
    \centering
    \includegraphics[width=\linewidth, height=0.9\textheight, keepaspectratio]{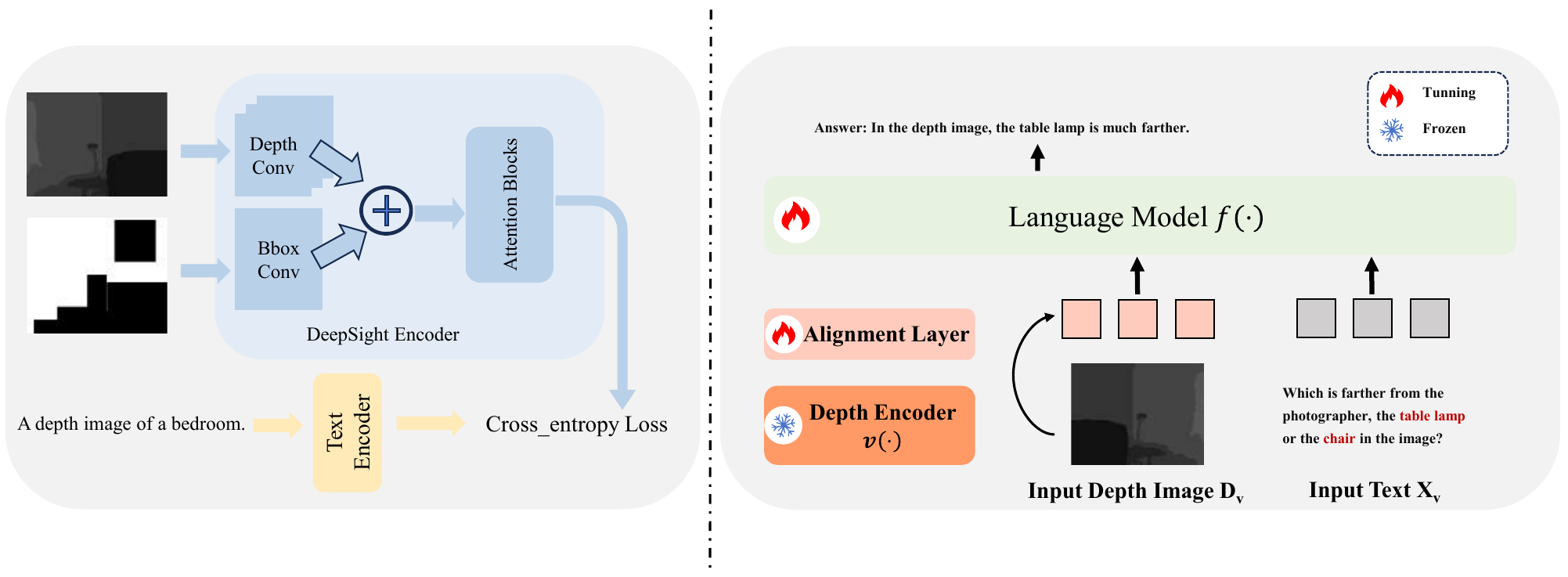}
    \caption{\textbf{Overall model architecture.} The left figure shows the Depth Encoder architecture, where DeepSight modifies the CLIP image encoder to take an additional Bbox channel along with depth convolution. The right figure illustrates the model pipeline, showing the data flow from the input question and depth image to the output answer. In the alignment stage, only the Alignment Layer is trained while keeping the Depth Encoder, Text Encoder, and LLM frozen. In the fine-tuning stage, the other modules remain frozen and additional training is performed on the LLM.}
    \label{fig:model}
\end{figure*}

\textbf{Caption Scoring.} In the COCO dataset, each image is annotated with multiple captions describing diverse scene aspects. However, to enhance the alignment between the image and its corresponding depth representation, we select only the caption that is most relevant to the depth image's semantics. We use the LanguageBind ~\cite{radford2021learningtransferablevisualmodels} Depth Encoder to compute similarity scores between each caption and the depth image. By computing similarity scores for all the captions, we identify the highest-scoring caption, ensuring that it is the one most semantically consistent with the depth image. This process helps to filter out captions that may not accurately reflect the depth-related content, ensuring that the chosen caption has a strong semantic match with the depth image. As a result, this approach guarantees a high level of semantic similarity between the depth image and the selected caption. Ultimately, we obtain a total of 118k high-quality depth-text-bboxes pairs, which will be used for subsequent model training.

\textbf{Instruction Generation.} We introduce the Depth Instruction Dataset, constructed in a format similar to LLaVA, to improve the alignment between the depth images and textual understanding. The dataset is generated by providing GPT-3.5 with a depth image's corresponding captions, and object bounding boxes. It is worth noting that for each image, we need to input multiple corresponding captions into GPT-3.5 to ensure that it can understand the image information from various perspectives. Leveraging this data, GPT-3.5 synthesizes instruction-following examples using predefined templates. Ultimately, we obtain three types of instruction data, totaling 22k instances: complex reasoning, multi-round dialogue, and detailed description.

\section{DeepSight Vision Encoder}
In this section, we introduce the vision encoder of DeepSight, a modified version of the original CLIP model that utilizes local object depth information to enhance its understanding of the entire depth image.

\subsection{Vision Encoder Architecture}
The vision encoder introduces structural modifications to the CLIP architecture to enhance its ability to recognize local information, while preserving the model's existing knowledge. These changes enable the model to better process depth features, which are essential for understanding 3D scenes. At the same time, it retains CLIP's strength in large-scale text-image pretraining, allowing it to effectively integrate spatial and semantic understanding for tasks that involve both depth and visual comprehension.

As shown in Figure~\ref{fig:model}, we add a bounding box convolution layer to the ViT architecture for local depth information recognition. 
Given the depth image \( D \) and the corresponding object bounding box mask \( D \), the CLIP patch division encoding is applied to both inputs. The encoded depth image \( D \) is processed by the Depth Conv, and the encoded bounding box mask \( D \) is processed by the Bbox Conv. The resulting feature representations are denoted as \( H_D = f(D) \) and \( H_M = g(M) \), respectively. These two results are then combined and passed into the subsequent attention module, which produces the final vision encoding result \(H_V = H_D + H_M \). It is important to note that the input to the Bbox Conv is a single-channel matrix, where the values are binary: \( 0 \) represents the background and \( 1 \) represents the object. The text input \( T \) is directly fed into the CLIP text encoder. The image embedding \(H_V\) and text embedding \(H_T\) are then used to compute a cross-entropy loss, aligning the multimodal representations to ensure coherence between the depth image and the textual description, after applying softmax processing to both embeddings. The cross-entropy formula is expressed as:
\[
\mathcal{L}_{\text{CE}} = - \sum_{i=1}^{N} \left[ H_V \log(\hat{H_T}) + H_T \log(\hat{H_V} ) \right]
\]
Where \(H_V\) and \(H_T\) are the image and text embeddings after softmax processing, and \( \hat{H_V} \) and \( \hat{H_T} \) are the predicted image and text embeddings.

During inference, depth convolution takes the depth image as input, and the bounding box convolution layer input defaults to 1.
\subsection{Training Method}
During training, we keep the CLIP text encoder frozen and only train the visual encoder. We employ a specific data sampling strategy where depth-box-text pairs are randomly replaced with depth-text pairs. This approach helps preserve the model's prior knowledge while maintaining its ability to understand the image at a global level. Initially, we set the sample parameter \(r=0.1\), , randomly replacing depth-box-text pairs with a ratio of \(r\). At the same time, the input to the bounding box convolution layer is set to 1 for all replacements. During inference, the input to the bounding box convolution layer is likewise set to 1.

\section{Aligning with LLM}
To the best of our knowledge, there are currently no multimodal large language models specifically designed for the depth modality. In this section, we introduce a large language model tailored for depth understanding, leveraging depth encoder and Vicuna alignment to enhance the model’s ability to process depth information ~\cite{wu2023qalignteachinglmmsvisual}.
\subsection{Model Architecture}

As shown in Figure ~\ref{fig:model}, the DeepSight model consists of three main components: a depth modality encoder, a trainable linear projection layer, and a large language model. Specifically, DeepSight is used as the depth encoder to process depth images, while an MLP (multi-layer perceptron) serves as the alignment layer to correlate the depth information with the text. Vicuna1.5-7B is used as the foundational language model to handle the combined embeddings from both text and depth images. The model takes text and depth images as inputs, aligning the depth features with the text via the MLP, and then merges both modalities' embeddings. These combined representations are fed into the language model, which generates the appropriate response or output. This architecture enhances the model’s ability to process and interpret multimodal data by integrating depth information with text, allowing for more accurate, context-aware responses that involve complex spatial reasoning and depth perception.
\subsection{Alignment}
Similar to LLaVA, we train the model in two stages to optimize its performance. In the first stage, called the alignment stage, we utilize 118k depth-text pairs from section 4 to train the linear projection layer. The purpose of this phase is to align the DeepSight encoder with the Vicuna1.5-7B model, ensuring that the depth image feature space and the text embedding space are effectively synchronized. This alignment is crucial because it allows the model to process and understand both modalities in a unified manner. During this stage, we freeze both the depth encoder and the large language model, focusing exclusively on training the linear projection layer. By doing so, we isolate the training process, allowing for a more precise adjustment of the mapping between the depth image features and the textual embeddings. This helps to ensure that the depth and text modalities are properly aligned, which is essential for the model’s ability to handle and interpret both types of data. Once this alignment is achieved, it sets a solid foundation for the subsequent stages of training, where the model can refine its understanding and improve its performance on more complex tasks.
\subsection{Supervised Fine-tuning}
The supervised fine-tuning (SFT) stage is designed to enhance the model’s ability to respond to depth-related instructions, building on the alignment established in the previous stage. During this phase, we leverage a dataset of 22k Depth Instruction Dataset to further train the model. Unlike the alignment stage, where only the linear projection layer was trained, in this stage, we freeze the depth encoder and focus on fine-tuning both the linear projection layer and the large language model. Additionally, we minimize the depth-conditioned text generation loss to ensure the model generates accurate and contextually appropriate responses based on the depth-related instructions. This fine-tuning process allows the model to better handle complex interactions involving depth information, ensuring more accurate and relevant outputs when given depth-specific tasks. By combining the depth instruction data with the previously learned embeddings, the model becomes increasingly adept at processing depth information.

\begin{table*}[t]
  \caption{Performance comparison of depth-aware multimodal models on the Depth Template Benchmark, covering four QA sub-tasks: Scene Classification, Recognition, Distance Judgment, and Security. We evaluate both zero-shot and fine-tuned settings using the Depth Instruction Dataset. LanguageBind is first aligned with Vicuna1.5-7B before fine-tuning. Best results in each column are highlighted in bold.}
  \label{tab:benchmark_comparison}
  \centering
  {
    \begin{tabular}{lccccc}
      \toprule
      \textbf{Model} & \textbf{Scene Classification} & \textbf{Recognition} & \textbf{Distance Judge} & \textbf{Security} & \textbf{Avg.} \\
      \midrule
      \multicolumn{6}{c}{\textbf{(a) Zero Shot}} \\
      \midrule
      PandaGPT-7B & 28.19 & 17.15 & 13.74 & 43.16 & 25.56 \\
      ImageBindLLM-7B ~\cite{han2023imagebind} & 58.23 & 23.06 & 28.74 & 22.68 & 33.18 \\
      \textbf{DeepSight-7B (Ours)} & 57.01 & 28.72 & 39.23 & 29.15 & 38.53 \\
      \midrule
      \multicolumn{6}{c}{\textbf{(b) Fine-tuning}} \\
      \midrule
      PandaGPT-7B-FT & 36.92 & 24.38 & 33.12 & 44.06 & 34.62 \\
      ImageBindLLM-7B-FT & 60.19 & 27.33 & 41.82 & 29.78 & 39.28 \\
      LanguageBind-Aligned-7B-FT & 62.31 & 36.08 & 57.92 & 37.84 & 48.54 \\
      \textbf{DeepSight-7B (Ours)} & \textbf{64.86} & \textbf{40.56} & \textbf{63.17} & \textbf{44.81} & \textbf{53.85} \\
      \bottomrule
    \end{tabular}
  }
\end{table*}

\section{Experiment}
In this section, we conduct comprehensive experiments to evaluate the effectiveness of our proposed DeepSight model. First, we assess the zero-shot scene classification performance of the DeepSight vision encoder, comparing it with a set of strong baseline vision models. Then, We also fine-tune other models using the Depth Instruction dataset and evaluate them on our benchmark.

In addition to benchmark comparisons, we also perform ablation studies to investigate the impact of key architectural components. By systematically removing or modifying elements such as the Bbox channel, and alignment layer, we quantify the contribution of each component to the final performance. We also conduct experiments to validate the effectiveness of our constructed Depth Instruction Dataset and the data sample strategy.

These experiments collectively validate the design choices in our model. The results demonstrate that DeepSight not only excels in zero-shot classification tasks but also shows strong generalization to downstream depth-aware question answering tasks, confirming the effectiveness and robustness of our approach.

\subsection{Zero-shot Scene Classification}
We compared our DeepSight vision encoder with other depth-aware visual encoders on the zero-shot scene classification task, and the results demonstrate that our model exhibits state-of-the-art performance. As shown in Table~\ref{tab:zero-shot cls}, our vision encoder achieves an accuracy of 67.0\% and 38.4\% on 1,000 randomly selected samples from the NYU-D and SUN-D datasets, respectively, surpassing ImageBind by 13.0\% and 3.3\%, and LanguageBind by 1.9\% and 0.7\%. These results demonstrate that the DeepSight depth encoder effectively captures holistic image semantics while exhibiting superior performance in leveraging depth information for scene classification. The substantial improvements over existing models further substantiate the effectiveness of our encoder architecture.

\begin{table}[ht]
  \caption{\textbf{Zero-shot scene classification} on 1,000 NYU-D and 1,000 SUN-D images. Our model achieves state-of-the-art accuracy.}
  \label{tab:zero-shot cls}
  \centering
  \renewcommand{\arraystretch}{1.0}  % 恢复紧凑行距
  \begin{tabular}{@{}lcc@{}}
    \toprule
    \textbf{Model} & \textbf{NYU-D (\%)} & \textbf{SUN-D (\%)} \\
    \midrule
    ImageBind     & 54.0       & 35.1       \\
    LanguageBind  & 65.1       & 36.7       \\
    \textbf{DeepSight-7B (Ours)}          & \textbf{67.0} & \textbf{38.4} \\
    \bottomrule
  \end{tabular}
\end{table}

\subsection{Benchmark Comparison with other Models}
To thoroughly evaluate our model’s capacity for understanding depth images, we employ the Depth Template Benchmark, introduced in Section~\ref{Depth Template Benchmark}, as the primary evaluation framework. This benchmark comprises multiple sub-tasks designed to test different aspects of depth reasoning. All evaluation metrics are based on accuracy to enable fair and interpretable comparisons across models.

We first evaluate several depth-related models on the benchmark in a zero-shot setting, as shown in Table ~\ref{tab:benchmark_comparison}(a). Without any fine-tuning, DeepSight achieves the best overall performance with an average score of 38.54\%, outperforming PandaGPT-7B with 25.56\% and ImageBindLLM-7B with 33.18\% in the zero-shot setting.. It shows clear advantages in recognition and distance judgment tasks, demonstrating stronger depth understanding ability. It is observed that the DeepSight exhibits lower performance in scene recognition compared to the zero-shot scene classification capability of the depth encoder. This discrepancy can be attributed to the difference in evaluation settings, the current task is framed as a textual question answering task, rather than direct classification.

To ensure a rigorous and fair evaluation, we also fine-tune all baseline models using the Depth Instruction Dataset introduced in Section~\ref{Depth Instruction Dataset}. For consistency, we align the depth encoder of LanguageBind with Vicuna-1.5-7B, which serves as the LLM backbone in our own model. This alignment ensures that differences in performance are attributed to architectural design rather than discrepancies in model scale or backbone capabilities. All models—including PandaGPT\_7B, ImageBindLLM\_7B and the aligned LanguageBind are trained using exactly the same data as our DeepSight model.

As reported in Table~\ref{tab:benchmark_comparison}(b), our model demonstrates consistently superior performance across all four evaluation tasks. It achieves accuracy rates of 64.86\% on Scene Classification, 40.56\% on Recognition, 63.17\% on Distance Judgment, and 44.81\% on Security Assessment. These results not only surpass those of all competing baselines but also further substantiate the robustness and generalization capability of our proposed architecture. In addition, both PandaGPT-7B and ImageBindLLM-7B-FT exhibit notable performance improvements after fine-tuning on our Depth Instruction Dataset, confirming the effectiveness of the constructed instructions in enhancing the depth-related understanding of vision models.

These substantial gains highlight the value of our proposed architecture and the effectiveness of incorporating depth-specific features into the model’s representation learning. The results validate the impact of our generated depth data in significantly enhancing the model’s ability to interpret and reason about depth-aware visual inputs. By leveraging structural and semantic cues uniquely available in depth images—further enhanced through Bbox-guided attention and instruction-aligned training—our model demonstrates strong reasoning capabilities and adaptability to complex multimodal scenarios. Collectively, these findings confirm that our model not only excels on current benchmarks but also establishes a new standard for depth-aware understanding in vision-language systems.

\subsection{Ablations}
We perform a series of ablation studies to better understand the contribution of key design choices in our framework. Specifically, we investigate the effects of the fine-tuning strategy, the inclusion of the Bbox Convolution layer, the use of our instruction dataset, and the data sampling strategy. These experiments are designed to isolate the role of each component and assess how they individually and collectively affect the model's overall performance.

\textbf{SFT Choice.} To verify the importance of fine-tuning the LLM during the SFT stage, we conduct an experiment in which we selectively train either the MLP or the LLM. The Distance Judge task is chosen as the evaluation task for this experiment. In this experiment, we compared the performance of three different training configurations: fine-tuning only the MLP, fine-tuning only the LLM and fine-tuning both the MLP and the LLM together. As shown in Table ~\ref{tab:ablation}, fine-tuning both the MLP and LLM together resulted in performance improvements of 16.46\% and 5.81\% compared to training only the MLP or the LLM, respectively. This demonstrates that jointly fine-tuning both the MLP and the LLM during the SFT stage is crucial for maximizing the model's performance, and highlights the importance of fine-tuning the LLM in conjunction with the alignment layer to fully leverage the depth information in the task. The significant performance gains reinforce the idea that the collaborative training of both components leads to a more effective model.

\begin{table}[ht]
  \caption{\textbf{Ablation study on SFT structure.} Ablation study on SFT structure. Results of different configurations evaluated on the Distance Judge sub-task of the Depth Template Benchmark.}
  \label{tab:ablation_sft}
  \centering
  \renewcommand{\arraystretch}{1.0}
  \begin{tabular}{@{}ccc@{}}
    \toprule
    \textbf{Train MLP} & \textbf{Train LLM} & \textbf{Distance Judge} \\
    \midrule
    \ding{51}  & \ding{55}  & 46.71 \\
    \ding{55}  & \ding{51}  & 57.36 \\
    \ding{51}  & \ding{51}  & \textbf{63.17} \\
    \bottomrule
  \end{tabular}
  \Description{Ablation table showing distance judgment accuracy when training only MLP, only LLM, or both components.}
\end{table}

\textbf{Bbox Convolution.} To evaluate the effectiveness of the Bbox Convolution layer in understanding depth information, we conduct an ablation study comparing the performance of distance judgment under different training and inference configurations. As shown in Table~\ref{tab:ablation}, when the Bbox Convolution layer is included during training but excluded during inference, the model achieves a distance judgment accuracy of 58.46\%. In contrast, when the Bbox Convolution layer is used during both training and inference, the accuracy improves to 63.17\%. These results indicate that retaining the Bbox Convolution layer throughout both stages significantly enhances the model’s ability to reason about depth. This highlights the crucial role of the Bbox Convolution layer in capturing depth-related features and improving performance on distance judgment tasks.

\begin{table}[ht]
  \caption{\textbf{Ablation on the Bbox Convolution.} Results of different configurations evaluated on the Distance Judge sub-task of the Depth Template Benchmark.}
  \label{tab:ablation}
  \centering
  \renewcommand{\arraystretch}{1.3}
  \begin{tabular}{@{}ccc@{}}
    \toprule
    \textbf{Train Bbox Conv} & \textbf{Infer Bbox Conv} & \textbf{Distance Judge} \\
    \midrule
    \ding{51} & \ding{55} & 58.46 \\
    \ding{51} & \ding{51} & \textbf{63.17} \\
    \bottomrule
  \end{tabular}
  \Description{Ablation showing the effect of training/inference with Bbox Convolution on distance judgment accuracy.}
\end{table}

\begin{figure}[ht]
    \centering
    \includegraphics[width=0.9\linewidth]{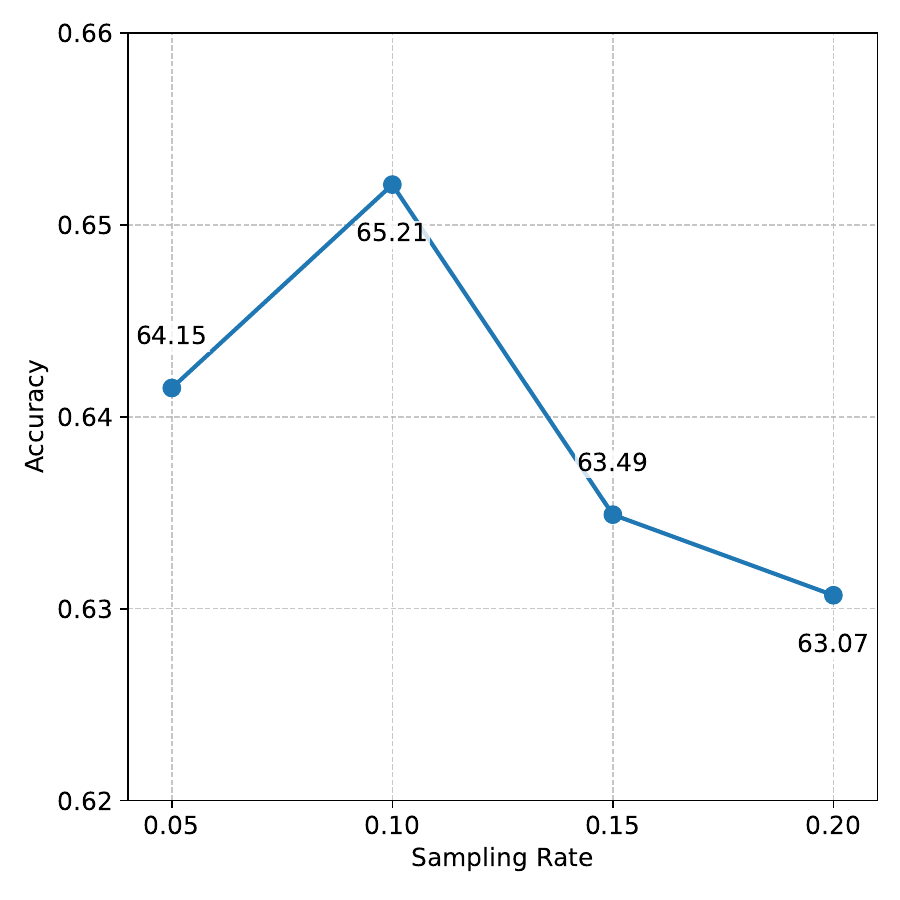}
    \caption{Sample ratio search experiment. We search sample ratio with a step of 0.05. Test metric is zero-shot Scene Classification top-1 accuracy.}
    \label{fig:sample}
\end{figure}

\textbf{Data Sample Strategy.} To preserve CLIP’s ability to understand the global structure of depth images, we randomly replace a portion of the depth-text-bboxes pairs with depth-text pairs during training. This strategy allows the model to maintain holistic scene understanding while also acquiring the ability to process localized, object-level information.
We use zero-shot Scene Classification top-1 accuracy as the evaluation metric. Specifically, we adopt the ViT-B/16 model and train it on the depth-text-bounding box dataset for 5 epochs. As illustrated in Figure~\ref{fig:sample}, the model without data sampling performs worse than those incorporating sampling. Moreover, we observe that overly high replacement ratios can degrade performance. Based on these observations, we adopt a sample ratio of 0.1 as the optimal setting.

\begin{table*}[t]
  \caption{Performance comparison of general vision-language models on the Depth Template Benchmark, covering four QA sub-tasks: Scene Classification, Recognition, Distance Judgment, and Security. We evaluate both zero-shot and fine-tuned settings using the Depth Instruction Dataset.}
  \label{tab:vision_model}
  \centering
  \begin{tabular}{lccccc}
    \toprule
    \textbf{Model} & \textbf{Scene Classification} & \textbf{Recognition} & \textbf{Distance Judge} & \textbf{Security} & \textbf{Avg.} \\
    \midrule
    \multicolumn{6}{c}{\textbf{(a) Zero-shot}} \\
    \midrule
    LLaVA-7B & 30.62 & 26.15 & 25.41 & 20.58 & 25.69 \\
    BLIP-Vicuna-7B & 24.17 & 29.41 & 30.14 & 31.07 & 28.20 \\
    QWen2.5-VL-3B ~\cite{bai2025qwen2} & 36.83 & 26.75 & 31.02 & 27.58 & 30.05 \\
    QWen2.5-VL-7B & \textbf{47.59} & \textbf{34.17} & \textbf{46.90} & \textbf{34.25} & \textbf{40.73} \\
    \midrule
    \multicolumn{6}{c}{\textbf{(b) Fine-tuning}} \\
    \midrule
    LLaVA-7B & 41.53 & 29.71 & 34.60 & 27.39 & 33.81 \\
    BLIP-Vicuna-7B & 47.25 & \textbf{40.96} & 49.15 & 38.26 & 43.91 \\
    QWen2.5-VL-3B & 43.16 & 30.44 & 42.97 & 31.17 & 36.44 \\
    QWen2.5-VL-7B & \textbf{60.62} & 39.71 & \textbf{58.36} & \textbf{40.60} & \textbf{49.32} \\
    \bottomrule
  \end{tabular}
\end{table*}

\textbf{Instruction Dataset} To validate the effectiveness of our constructed Depth Instruction Dataset in enhancing the stereoscopic perception capabilities of multimodal vision models, we conduct additional fine-tuning on several representative vision-language models and evaluate their performance using the \textit{Depth Template Benchmark}. To ensure compatibility, the depth images are converted to three-channel format and directly used as input across all models, maintaining consistency in training and evaluation settings.

As shown in Table~\ref{tab:vision_model}, all evaluated models demonstrate significant performance improvements after fine-tuning with our dataset. For example, LLaVA-7B shows an average accuracy gain from 25.69\% to 33.81\%, while BLIP-Vicuna-7B achieves a substantial boost from 28.20\% to 43.91\%. Similarly, both variants of QWen2.5-VL benefit from our instruction data, with the 7B model attaining the highest overall performance, increasing from 40.73\% to 49.32\%. These results not only confirm the value of task-aligned supervision with depth-specific semantics, but also verify that traditional vision-language models, when equipped with explicit depth cues, can better understand the stereoscopic structure and spatial layout of visual scenes.

\section{Case Study}
\begin{figure}[ht]
    \centering
    \includegraphics[width=0.9\linewidth]{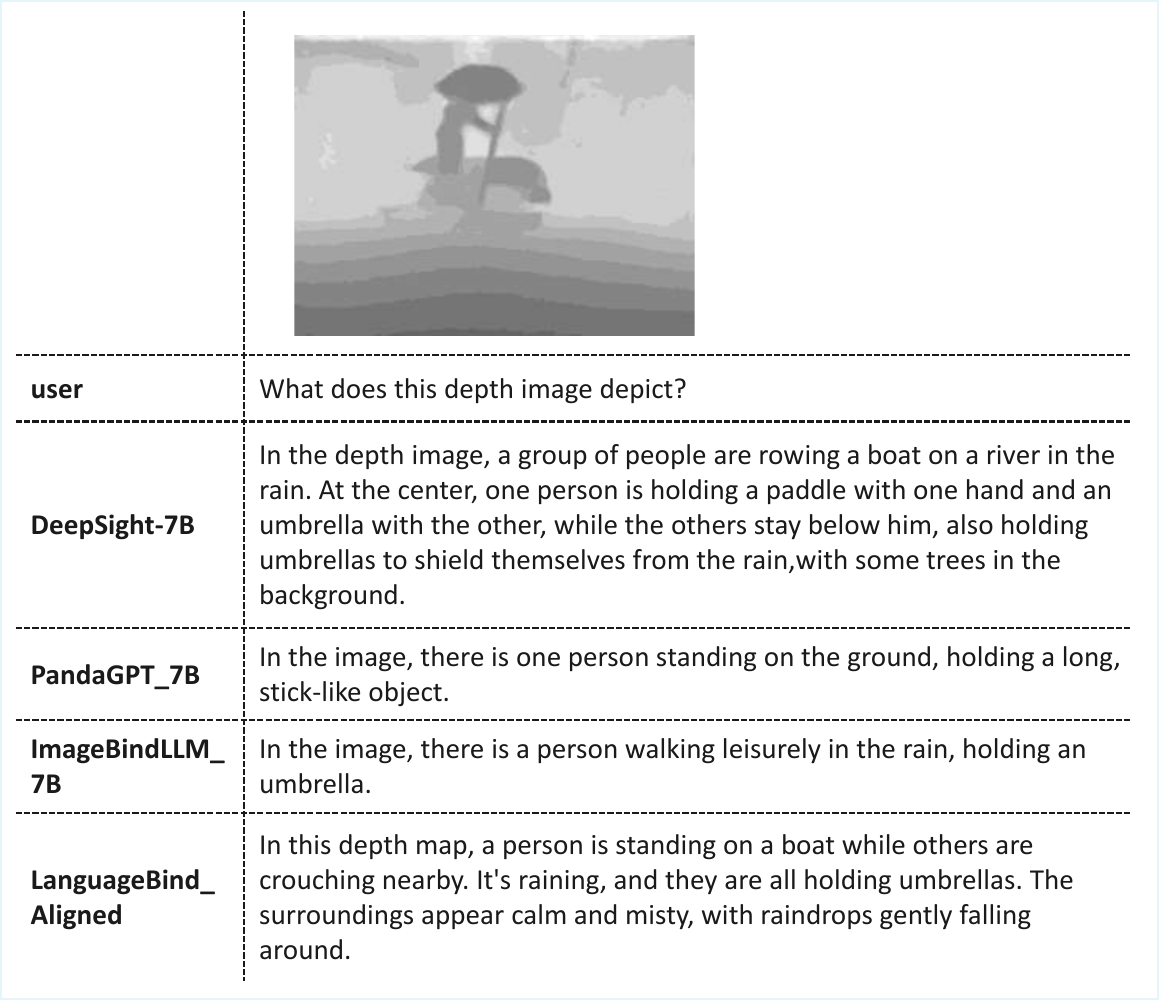}
    \caption{Case study. We posed questions about the depth image to multiple models, and the results show that DeepSight is better able to meet the task requirements.}
    \label{fig:case}
\end{figure}

To further assess the effectiveness of our model, we present a case study on representative examples. As shown in Figure~\ref{fig:case}, DeepSight captures both global scene context and fine-grained object details, demonstrating a balanced understanding of spatial layout and semantics. In contrast, other models show issues such as incorrect object recognition or color description. This comparison highlights how incorporating explicit depth cues improves multimodal models’ fine-grained and context-aware visual understanding, and underscores the value of jointly reasoning over global structure and local details.

This case study further supports our claim that explicitly incorporating depth information, as done in DeepSight, improves multimodal models' ability to understand visual scenes both globally and locally. Depth helps the model grasp spatial relationships, object positioning, and occlusions, enabling more precise and context-aware reasoning. It also underscores the importance of designing architectures that can jointly reason about overall scene layout and fine-grained semantics.

\section{Conclusion}
In this paper, we present DeepSight, a depth-modality-specific multimodal large language model designed to enhance the understanding of depth images in multimodal tasks. We first construct the QA Depth Template Benchmark based on real-world datasets to evaluate both global and local depth reasoning. To support model training, we convert the COCO dataset into depth images, generating 118k depth-text pairs and 22k instruction samples.

We further propose DeepSight, a modified CLIP architecture aligned with large language models through fine-tuning. Finally, ablation experiments validate the effectiveness of both our architectural design and training strategy.Our work highlights the potential of depth-aware multimodal models in advancing stereoscopic visual understanding.

\bibliographystyle{ACM-Reference-Format}
\bibliography{sample-base}

\end{document}